\documentclass[10pt,twocolumn,letterpaper]{article}

\usepackage{wacv}              
\usepackage[numbers]{natbib}
\usepackage{cuted}
\usepackage{capt-of}
\usepackage{amsmath} 
\usepackage[table,xcdraw]{xcolor}
\usepackage{multirow}
\usepackage{amssymb}
\usepackage{float}
\usepackage{color, soul}
\usepackage{enumitem}
\usepackage{graphicx}
\usepackage{subcaption}
\usepackage{booktabs}
\usepackage{subcaption}
\usepackage{multirow}
\usepackage{dirtytalk}
\usepackage{multirow}
\usepackage[table,xcdraw]{xcolor}
\usepackage[utf8]{inputenc}
\newcommand{\red}[1]{\textcolor{black}{#1}}

\usepackage{adjustbox}
\usepackage{graphicx}
\usepackage{subcaption}
\usepackage{adjustbox}
\usepackage{pgffor}
\usepackage{graphicx}
\usepackage{flafter}
\usepackage[section]{placeins}
\usepackage{makecell}
\usepackage{currfile}
\usepackage{ulem}
\usepackage{tabularx}
\DeclareUnicodeCharacter{2009}{\,}
\definecolor{wacvblue}{rgb}{0.21,0.49,0.74}
\usepackage[pagebackref,breaklinks,colorlinks,allcolors=wacvblue]{hyperref}

\def\wacvPaperID{240} 
\def\confName{WACV}
\def\confYear{2026}

\title{Efficient Text-Guided Convolutional Adapter for the Diffusion Model}

\author{
Aryan Das\\
VIT Bhopal\\
{\tt\small aryan.das2021@vitbhopal.ac.in}
\and
Koushik Biswas\\
IIIT Delhi\\
{\tt\small koushikb@iiitd.ac.in}
\and
Swalpa Kumar Roy\\
Tezpur University Assam\\
{\tt\small swalpa@tezu.ernet.in}
\and
Badri Narayana Patro\\
IIT Kanpur\\
{\tt\small badri@iitk.ac.in}
\and
Vinay Kumar Verma\\
IIT Kanpur\\
{\tt\small vinayugc@gmail.com}
}

\begin{document}

\maketitle
\begin{abstract}
We introduce the \textbf{Nexus Adapters}, novel text-guided efficient adapters to the diffusion-based framework for the Structure Preserving Conditional Generation (SPCG). Recently, structure-preserving methods have achieved promising results in conditional image generation by using a base model for prompt conditioning and an adapter for structure input, such as sketches or depth maps. These approaches are highly inefficient and sometimes require equal parameters in the adapter compared to the base architecture. It is not always possible to train the model since the diffusion model is itself costly, and doubling the parameter is highly inefficient. In these approaches, the adapter is not aware of the input prompt; therefore, it is optimal only for the structural input but not for the input prompt. To overcome the above challenges, we proposed two efficient adapters, \textbf{Nexus Prime and Slim}, which are guided by prompts and structural inputs. Each Nexus Block incorporates cross-attention mechanisms to enable rich multimodal conditioning. Therefore, the proposed adapter has a better understanding of the input prompt while preserving the structure. \red{We conducted extensive experiments on the proposed models and demonstrated that the Nexus Prime adapter significantly enhances performance, requiring only 8M additional parameters compared to the baseline, T2I-Adapter. Furthermore, we also introduced a lightweight Nexus Slim adapter with 18M fewer parameters than the T2I-Adapter, which still achieved state-of-the-art results}. Code: \href{https://github.com/arya-domain/Nexus-Adapters}{https://github.com/arya-domain/Nexus-Adapters}
\end{abstract}
\section{Introduction}\label{sec:intro}

Text-to-image (T2I) generation has advanced very fast in recent years, mainly due to the success of large diffusion models trained on diverse and extensive image-text datasets. However, Stable Diffusion~\cite{stablediffusion} has become popular by using Denoising Diffusion Probabilistic Models (DDPMs) in conjunction with pretrained vision-language models such as CLIP to generate semantically rich and visually coherent images from natural language prompts. However, while these models are good at creating images from text, they often have trouble following exact layouts, structures, or small visual details. This limitation becomes especially problematic in real-world applications where user intent often combines high-level semantics with low-level control signals, such as edge maps, segmentation masks, or human poses.

To address this, several recent works have proposed augmentations to the diffusion pipeline that incorporate additional conditioning mechanisms. ControlNet~\cite{controlnet} proposed a dual-branch architecture that introduces trainable control networks alongside the frozen diffusion backbone and that enables the guidance of external signals such as depth maps or line drawings. In a similar vein, T2I-Adapter~\cite{t2i} proposes lightweight adapter modules that are trained to inject structural information into the frozen backbone, offering a plug-and-play solution that supports various structural conditions without retraining the base model. ControlNet++~\cite{controlnet++} further refines the ControlNet concept by introducing a cyclic consistency loss to ensure alignment between the generated image and the input structure. Meanwhile, CtrLoRA~\cite{ctrlora} proposes a parameter-efficient method by using low-rank adaptation (LoRA) layers to condition the backbone with minimal additional overhead, making it suitable for resource-constrained scenarios.

Despite these advances, current methodologies continue to exhibit several limitations. First, many approaches necessitate direct modifications to the primary diffusion backbone or involve extensive fine-tuning, thereby undermining generalizability and increasing training complexity. Second, parameter efficiency remains a critical challenge. Architectures like ControlNet and ControlNet++ demand nearly as many parameters as the base model encoder, making them highly inefficient for large-scale models. Even methods such as CtrLoRA rely heavily on ControlNet as their adapter backbone, limiting true efficiency gains. Finally, the adapter operates independently of the given prompt and only receives structural signals as input, as in the T2I Adapter. Consequently, the model aims to preserve structural information without leveraging the semantic context of the prompt. Incorporating prior knowledge of the input prompt into the adapter may provide stronger guidance and facilitate more effective task adaptation.

\red{
To address the above challenges, this study introduces the Nexus Adapters, prompt-guided and parameter-efficient adapters with the following key contributions:
\textit{(1)~Prompt-Driven Conditional Guidance:} To our best knowledge, we are the first to propose conditional-focused alignment within the adapter, jointly attending to prompt and structural cues for stronger guidance.
\textit{(2)~Efficient Architecture:} We propose an effective convolutional design with cross-attention, which delivers high performance while using significantly fewer parameters.
\textit{(3)~State of the Art Performance:} Extensive experiments across diverse conditional tasks show that the Nexus Adapter parallels or surpasses recent baselines while remaining the most efficient.
}


\vspace{-3mm}
\section{Related Work}
\textbf{Image Synthesis and Conditional Generation:}
Early image synthesis methods faced significant challenges due to the high dimensionality and structural complexity of natural images. Generative Adversarial Networks (GANs)~\cite{gan}, Variational Autoencoders (VAEs)~\cite{vae}, and flow-based models~\cite{flow} have been proposed to enable realistic image generation through efficient sampling and stable optimization processes. Initial works in this domain focused largely on unconditional generation. However, conditional image synthesis soon gained attention, where external information such as semantic maps, sketches, or textual descriptions served as generation guides. Conditional GANs~\cite{i2i1,i2i2,i2i3,i2i4} enabled domain-to-domain translation by learning mappings from structural inputs to natural image outputs. Concurrently, T2I generation emerged~\cite{t2i1,t2i2,t2i3,t2i4}, aiming to synthesize images based on descriptive prompts. These approaches often treated each condition independently, lacking a unified and flexible mechanism to handle multimodal conditioning effectively~\cite{mul1}.

\textbf{Diffusion Models and Text-to-Image Synthesis:}
Diffusion models~\cite{diff} have recently gained traction for their strong generative capabilities, particularly in T2I tasks. These models generate images by progressively denoising a Gaussian noise input through a learned reverse diffusion process~\cite{phy1,phy2}. Notable models like GLIDE~\cite{glid}, DALL-E~\cite{t2i2}, Imagen~\cite{t2i1}, and Stable Diffusion~\cite{ldm} integrate text features into denoising blocks using cross-attention mechanisms. Although these models achieve impressive visual fidelity, they often fail to capture fine-grained structure or maintain spatial consistency, especially when guided by ambiguous or complex prompts. To address these shortcomings, structure-preserving methods such as PITI~\cite{piti} and other editing-based approaches~\cite{p2p,edit1,edit2} attempt to close the gap between text semantics and structural features through optimization, attention manipulation, or guidance gradients. However, these methods often suffer from inefficiency, as they either require inference-time optimization or lack direct parameter learning.

\textbf{Adapters for Controllable Diffusion:}
In pursuit of controllable T2I generation, recent works have introduced modular extensions to frozen diffusion backbones. ControlNet~\cite{controlnet} and T2I-Adapter~\cite{t2i} inject trainable modules for conditioning the generation process on visual structure such as depth, edges, or segmentation maps. While these methods significantly enhance controllability, they often suffer from high parameter overhead, with adapters sometimes matching or exceeding the base model’s size. Moreover, these adapters are generally agnostic to the input text prompt, limiting their ability to effectively bridge visual and linguistic modalities. To mitigate these limitations, our work introduces Nexus Adapters—lightweight convolutional modules that incorporate prompt-aware cross-attention within each Nexus Block. In contrast to existing models, Nexus Adapters maintain structure fidelity while being guided semantically by the text input, leading to improved multimodal alignment.

\textbf{Parameter Efficiency and Prompt Awareness:}
A key limitation in existing structure-preserving conditional generation approaches lies in their parameter inefficiency. Models such as ControlNet~\cite{controlnet} and T2I-Adapter~\cite{t2i} introduce substantial overhead by training large, condition-specific adapters that often double the parameter count relative to the base diffusion model. Additionally, these models generally treat visual control inputs and text prompts independently, failing to leverage the semantic synergy between them during feature modulation. 

\red{\textbf{Generative Models on Image Editing:}
Generative models, particularly diffusion transformers, have emerged as powerful tools for image editing, surpassing traditional approaches by enabling controllable and high-quality transformations. Recent works have introduced benchmarks such as IDEA-Bench~\cite{idea} for professional-level evaluation, enabled real-time video customization through diffusion models~\cite{videomaker}, and developed training-free frameworks for task-agnostic applications~\cite{chatdit}. Advances in in-context learning and LoRA~\cite{loradit}, multitask unsupervised systems~\cite{gdt}, and instructional editing techniques~\cite{icedit} have further broadened the scope of generative editing. Moreover, context-driven insertion methods allow seamless integration of new elements into images~\cite{insertanything}, strengthening practical adoption in multimedia design.}

\vspace{-3mm}
\section{Methodology}
\begin{figure*}[t]
    \centering
    \vspace{-5mm}
    \includegraphics[width=0.75\linewidth, height=0.34\linewidth]{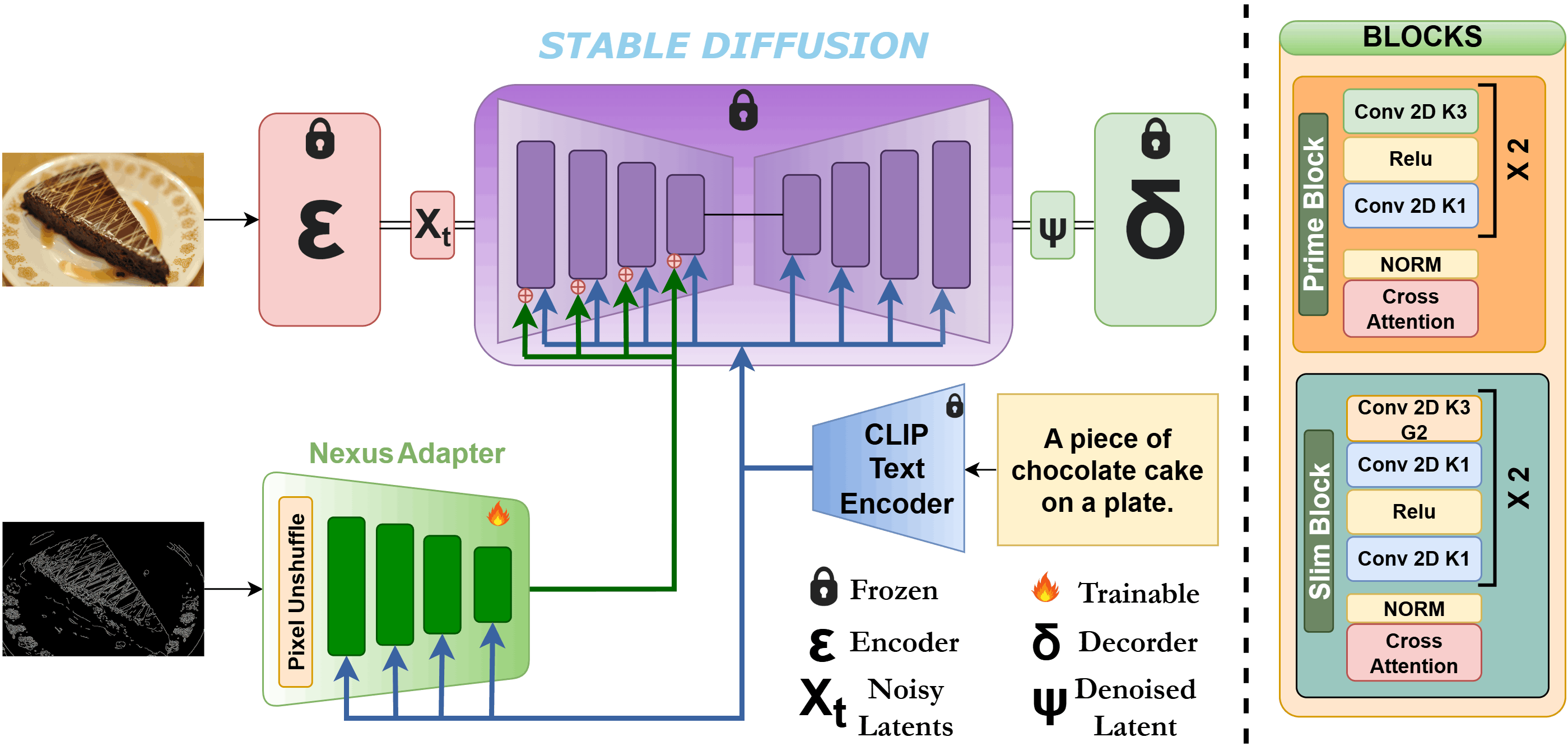}
    \vspace{-2mm}
    \caption{Architecture of our method. The framework consists of two parts: (1) A frozen Stable Diffusion model. (2) A lightweight Nexus Adapter trained to inject external condition signals. A condition image is processed through the adapter to produce multiscale features, while a text prompt is encoded by a frozen CLIP Text Encoder. Both visual and textual features are injected into the UNet via additive fusion and cross-attention. Detailed designs of Prime and Slim Nexus Blocks are shown on the right.}
    \label{fig:proposed_model}
\vspace{-5mm}
\end{figure*}

\subsection{Background: Diffusion Models}
Diffusion models gradually add noise to data in a forward process and then learn to reverse this noising to generate new samples. Formally, given a clean image $\mathbf{X}_{0}$, we define a Markov chain of noisy latents:
\begin{equation}
q(\mathbf{X}_{t} \mid \mathbf{X}_{t-1}) 
= \mathcal{N}\bigl(\mathbf{X}_{t}; \sqrt{1-\beta_{t}}\;\mathbf{X}_{t-1},\;\beta_{t}\mathbf{I}\bigr),
\end{equation}
where $\{\beta_{t}\}_{t=1}^{T}$ is a fixed schedule of variances.  After $T$ steps we obtain nearly pure noise. The reverse (denoising) process is modeled as:
\begin{equation}
p_{\theta}(\mathbf{X}_{t-1} \mid \mathbf{X}_{t})
= \mathcal{N}\bigl(\mathbf{X}_{t-1};\;\mu_{\theta}(\mathbf{X}_{t},t),\;\Sigma_{\theta}(t)\bigr),
\end{equation}
where the network $\mu_{\theta}$ learns to predict the original image at each timestep.  In practice, one often parameterizes $\mu_{\theta}$ via a noise-prediction network $\epsilon_{\theta}$, yielding the well‑known loss
\begin{equation}
\mathcal{L}(\theta)
= \mathbb{E}_{t,\mathbf{X}_{0},\epsilon}\Bigl[\big\|\epsilon - \epsilon_{\theta}(\mathbf{X}_{t},t)\big\|^{2}\Bigr].
\end{equation}

We utilize a pretrained Latent Diffusion Model (LDM), identical in architecture to Stable Diffusion (SD)~\cite{stablediffusion}, as the generative backbone. SD is a two-stage model comprising a convolutional autoencoder and a denoising UNet. The autoencoder first compresses an input image $\mathbf{X}_{0}$ into a latent representation $\mathbf{X}_{0} = \varepsilon(\mathbf{X}_0)$ (note that for the simplicity we represent the original and latent image by $\mathbf{X}_{0}$) using the encoder $\varepsilon$, and reconstructs it via the decoder $\delta$ as $\hat{x} = \delta(\Psi)$, where $\Psi$ is the denoised latent. During training, a forward diffusion process incrementally perturbs $\mathbf{X}_{0}$ over $T$ steps, producing noisy latent representations $\mathbf{X}_{t}$. The reverse denoising process is handled by a frozen UNet parameterized by $\theta$, which estimates the clean latent from noisy input:
\begin{equation}
\mathbf{X}_{t-1} = f_{\theta}(\mathbf{X}_{t}, t, c),
\end{equation}
where $c$ denotes text conditioning features obtained via a pretrained CLIP encoder. The model’s cross-attention layers integrate $c$ as key-value inputs at every denoising step.




\subsection{Proposed Framework Overview}
Building on a pretrained, frozen Stable Diffusion backbone that encodes rich image priors and text conditioning via a CLIP encoder, the proposed approach introduces a lightweight \textit{Nexus Adapter} for conditional guidance as shown in Figure~\ref{fig:proposed_model}. Two adapter variants are designed: \textit{Nexus Prime}, using high-capacity convolutional blocks, and \textit{Nexus Slim}, employing grouped and $1 \times 1$ convolutions for parameter efficiency. Both variants integrate external structural cues and textual prompts, which are processed by the frozen CLIP encoder and injected through cross-attention within Nexus blocks. This architecture enables effective modulation of the frozen diffusion prior without altering pretrained weights, achieving adaptable and efficient conditional generation. This module serves two functions:
\vspace{-1mm}
\begin{itemize}[leftmargin=4pt]
  \item \textbf{Structural guidance}: It accepts auxiliary inputs such as edge maps or segmentation masks, processes them through a compact “adapter” network, and generates multi‐scale feature maps. The adapter is highly efficient, requiring only $\sim8M$ additional parameters to the base model. It employs a hybrid design combining group-wise and point-wise efficient convolutional architectures, significantly reducing the parameter count. \vspace{-1mm}

  \item \textbf{Semantic alignment}: It reuses the frozen CLIP text embeddings and re-injects them into both the encoder and decoder of the diffusion UNet via cross‐attention. In contrast to previous approaches, we also guide the Nexus Adapter using the input prompt through an additional cross-attention mechanism.
\end{itemize}

\subsection{Nexus Adapter}
The \textit{Nexus Adapter} is a lightweight auxiliary module designed to introduce task-specific conditioning into the generative process while preserving the pretrained denoising prior. Operating in parallel with the denoising UNet, the adapter introduces no modifications to the frozen backbone and instead injects guidance signals through additive feature fusion.

Let the condition image be denoted as \( I_c \in \mathbb{R}^{3 \times H \times W} \), and its corresponding textual prompt as \( P \). The prompt \( P \) is embedded using a pretrained and frozen CLIP text encoder \( \mathcal{T}_{\text{CLIP}}(\cdot) \), yielding a global semantic embedding \( \mathcal{T} = \mathcal{T}_{\text{CLIP}}(P) \in \mathbb{R}^{d} \), where \( d \) is the embedding dimension.

To process the condition image \( I_c \), the adapter first applies a pixel unshuffle operation to downsample its spatial resolution to \( 64 \times 64 \), resulting in an initial tensor \( \mathbf{X}_0^L \in \mathbb{R}^{C_0 \times 64 \times 64} \). This representation is then fed into a series of \( K = 4 \) hierarchical transformation blocks for multiscale feature extraction:
\begin{equation}
\mathbf{E}^k = \mathcal{A}^k(\mathbf{X}_{k-1}^L), \quad \text{for } k = 1, 2, 3, 4,
\end{equation}
where \( \mathcal{A}^k(\cdot) \) denotes the \( k \)-th transformation block, parameterized differently for Prime ($P$) and Slim ($S$) blocks.

Between each stage, spatial downsampling is performed using a strided convolution operation \( \phi(\cdot) \):
\begin{equation}
\mathbf{X}_k^L = \phi(\mathbf{E}^k), \quad \text{for } k < 4.
\end{equation}

At each stage \( k \), the output feature map \( \mathbf{X}_k \in \mathbb{R}^{C_k \times H_k \times W_k} \) is progressively transformed by reducing its spatial resolution, with a structured adjustment in channel dimensionality. In the first three stages (\( k = 1, 2, 3 \)), the height and width are sequentially halved, while the channel dimension is doubled relative to the base channel size \( C_0 \):
\[
C_k = 2 \cdot C_{k-1}, \quad H_k = \frac{1}{2} H_{k-1}, \quad W_k = \frac{1}{2} W_{k-1}.
\]
In the final stage (\( k = 4 \)), the spatial dimensions continue to shrink by a factor of 2, but the channel dimension reverts to the base size:
\[
C_4 = C_3, \quad H_4 = \frac{1}{2} H_3, \quad W_4 = \frac{1}{2} W_3.
\]
This configuration ensures a rich intermediate representation with expanded channel capacity in the early stages for enhanced feature learning, followed by dimensional compression at the final stage to harmonize with downstream fusion modules or output heads.


\subsection{Nexus Blocks}
Each \textit{Nexus Block} is a compact neural unit designed to refine the adapter feature representation while integrating semantic guidance from the text prompt. Two variants are employed: the \textit{Prime Block} and the \textit{Slim Block}, which differ in their convolutional structure and computational complexity. Both block types receive an input feature tensor \( \mathbf{X} \in \mathbb{R}^{C \times H \times W} \), along with a text embedding \( \mathbf{T} \in \mathbb{R}^{n \times d} \) derived from the frozen CLIP text encoder. However the previous adapter considers only the image as condition.

Each Nexus Block processes the input feature \( \mathbf{X} \) through two sequential transformation units, followed by a normalization step. 
The refined visual feature is then passed through a cross-attention mechanism, where the query is derived from the processed visual feature and the key-value pairs originate from the text embedding \( \mathbf{T} \). This facilitates semantic alignment between the visual condition features and the prompt content. This design allows each Nexus Block to adaptively modulate spatial features based on prompt semantics while preserving spatial locality. 

\textbf{Cross attention:} The cross-attention mechanism used in Nexus Blocks follows the same formulation as the one employed in Stable Diffusion. Formally, given the normalized feature map \( \mathbf{F}_n \in \mathbb{R}^{C \times H \times W} \), it is first reshaped into a sequence of tokens \( \mathbf{Q} \in \mathbb{R}^{m \times d} \), where \( m = H \times W \), to serve as the queries. The text embedding \( \mathbf{T} \in \mathbb{R}^{n \times d} \) is projected to form the keys \( \mathbf{K} \) and values \( \mathbf{V} \). The cross-attention output is computed as:
\vspace{-2mm}
\begin{align}
    \textit{CrossAttention}(\mathbf{Q}, \mathbf{K}, \mathbf{V}) = \textit{SoftMax}\left(\frac{\mathbf{Q} \mathbf{K}^\top}{\sqrt{d}}\right) \cdot \mathbf{V}
    \vspace{-2mm}
\end{align}
where \( d \) is the dimensionality of the embeddings. The attention output is then reshaped back to the original spatial dimensions and fused with the input via a residual connection. This formulation enables effective modulation of spatial features using the semantic content of the prompt, ensuring consistency and semantic alignment across the adapter and backbone features.


\subsubsection{Prime Block}\label{sec:prime_block}
The \textit{Prime Block} ($P$) is designed to refine the adapter feature using standard convolutional layers and integrate semantic context via cross-attention. Let the input feature be denoted by \( \mathbf{X} \in \mathbb{R}^{C \times H \times W} \), and the associated text embedding from the frozen CLIP encoder be \( \mathbf{t} \in \mathbb{R}^{n \times d} \), where \( n \) is the number of tokens and \( d \) is the embedding dimension. This block applies two sequential convolutional sub-blocks, each consisting of a \(3 \times 3\) convolution followed by ReLU and a \(1 \times 1\) convolution. The operations are defined as follows:
{\small
\begin{align}
\mathbf{P}_1 &= \mathrm{ReLU}\big(\mathrm{Conv}_{3\times3}(\mathbf{X})\big); \quad
\mathbf{P}_2 = \mathrm{Conv}_{1\times1}(\mathbf{P}_1) \\
\mathbf{P}_3 &= \mathrm{ReLU}\big(\mathrm{Conv}_{3\times3}(\mathbf{P}_2)\big); \quad 
\mathbf{P}_4 = \mathrm{Conv}_{1\times1}(\mathbf{P}_3)
\end{align}
}

The output feature is then normalized to stabilize its statistical distribution: 
\begin{align}
\mathbf{F}_{{P_n}} &= \textit{Norm}(\mathbf{P}_4)
\end{align}
In the earlier approach, the adapter is not aware of the input prompt. To handle the above, we incorporate semantic alignment. The normalized visual feature \( \mathbf{F}_{\text{n}} \) is passed through a cross-attention layer, where it attends to the text embedding \( \mathbf{T} \). This is formulated as:
\begin{align}
&\mathbf{F}_{P_{attn}} = \textit{CrossAttention}(\mathbf{F}_{P_n}, \mathbf{T}, \mathbf{T}) \\
&\mathbf{F}_{\text{P}} = \mathbf{F}_{P_n} + \mathbf{F}_{P_{attn}}
\end{align}

Here, the attention output is added back to the original normalized feature to preserve residual learning. This structure allows the Prime Block to adaptively enhance spatial features in the adapter pathway with textual context, facilitating semantically coherent conditioning.

\subsubsection{Slim Block}\label{sec:slim_block}
The \textit{Slim Block} ($S$) is a parameter-efficient variant of the Nexus Block that achieves significant computational savings through the use of depthwise convolutions. It is designed to refine adapter features while incorporating semantic guidance via cross-attention. Let the input feature and corresponding text embedding are $\mathbf{X}$ and $\mathbf{T}$ respectivelly.

To reduce the parameter count while preserving spatial expressiveness, the Slim Block replaces standard convolutions with depthwise convolutions. Specifically, \( \text{Conv}_{3 \times 3}^D \) denotes a depthwise \( 3 \times 3 \) convolution with a group size of 2. This design significantly lowers the number of learnable parameters compared to conventional dense convolutions. Each Slim Block consists of two sequential transformation units. Each unit comprises a depthwise convolution \( \text{Conv}_{3 \times 3}^D \), followed by a pointwise \(1 \times 1\) convolution, a ReLU activation, and another pointwise convolution. The operations are defined as:
\begin{align}
\mathbf{S}_1 &= \textit{Conv}_{3 \times 3}^D(\mathbf{X}); \quad
\mathbf{S}_2 = \textit{Conv}_{1 \times 1}(\mathbf{S}_1) \\
\mathbf{S}_3 &= \textit{ReLU}\left(\ \mathbf{S}_1 + \mathbf{S}_2\right); \quad
\mathbf{S}_4 = \textit{Conv}_{1 \times 1}(\mathbf{S}_3) \\
\mathbf{S}_5 &= \textit{ReLU}\left(\textit{Conv}_{1 \times 1}(\mathbf{S}_4)\right); \quad 
\mathbf{S}_6 = \textit{Conv}_{1 \times 1}(\mathbf{S}_5)
\end{align}

The resulting output is normalized to stabilize training:
\begin{align}
\mathbf{F}_{S_n} &= \textit{Norm}(\mathbf{S}_6)
\end{align}

To inject semantic information, the normalized feature \( \mathbf{F}_{S_n} \) attends to the CLIP-derived text embedding \( \mathbf{T} \) using a cross-attention mechanism:
\begin{align}
&\mathbf{F}_{S_{attn}} = \textit{CrossAttention}(\mathbf{F}_{S_n}, \mathbf{T}, \mathbf{T}) \\
&\mathbf{F}_{\text{S}} = \mathbf{F}_{S_n} + \mathbf{F}_{S_{attn}}
\end{align}

This residual formulation allows the Slim Block to maintain effective feature propagation while incorporating prompt-level semantics, all within a reduced parameter budget due to the use of \( \text{Conv}_{3 \times 3}^D \) operations.

\subsection{Fusion}
For each adapter stream $A_k \in \{\mathbf{P}, \mathbf{S}\}$ (i.e., Prime or Slim), the output consists of a set of condition features \( \mathbf{E}_k = \{\mathbf{E}^1, \mathbf{E}^2, \mathbf{E}^3, \mathbf{E}^4\} \), where \( \mathbf{E}_k \in \{\mathbf{F_P}, \mathbf{F_S}\} \). These features are aligned in both spatial resolution and channel dimension with the intermediate UNet encoder activations \( \mathbf{U}_k = \{\mathbf{U}^1, \mathbf{U}^2, \mathbf{U}^3, \mathbf{U}^4\} \). The corresponding condition and encoder features are fused via element-wise addition:
\begin{equation}
\hat{\mathbf{U}}^k = \mathbf{U}^k + \mathbf{E}^k, \quad \text{for } k \in \{1, 2, 3, 4\}.
\end{equation}

During this process, the frozen CLIP textual embedding \( \mathbf{t} \) is incorporated into the adapter blocks using cross-attention mechanisms. This enables the visual features to be modulated dynamically based on the semantic content of the input prompt. Architectural details of the adapter transformation blocks \( \mathcal{A}_k \) are provided in Sections~\ref{sec:prime_block} and~\ref{sec:slim_block}.

\section{Experiment}
We conducted extensive experiments over various diverse tasks and evaluated the model performance against the recent strong baselines. 

\textbf{Dataset:} We evaluate the proposed architecture on four conditioning modalities: \textit{Canny}, \textit{Depth}, \textit{Sketch}, and \textit{Segmentation}, using the COCO 2017 dataset~\cite{coco_dataset} with $\sim$164k training images. Each modality is trained independently to assess control-guided generation under varied structural inputs. For \textbf{Segmentation}, we use COCO-Stuff annotations; for \textbf{Depth}, maps are generated via MiDaS~\cite{midas}; for \textbf{Sketch}, edge maps are extracted using an edge prediction model~\cite{edge}; and for \textbf{Canny}, standard Canny edge detection is applied. Further training details are in the Supplementary.

\textbf{Evaluation Protocol:} Quantitative evaluation is conducted on the COCO validation set comprising $5k$ samples. We report Fréchet Inception Distance (FID)~\cite{fid} to measure visual fidelity and CLIP Score~\cite{clips} (using ViT-L/14) to assess semantic alignment between generated images and the input prompts.
All models utilize the Stable Diffusion v1.5 backbone with frozen weights.

\section{Results and Discussion}
The section evaluates the quantitative, qualitative, and ablation study results for the various proposed components.

\subsection{Quantitative Comparison}\label{QC}

\begin{figure*}[t]
    \vspace{-5mm}
    \centering
    \includegraphics[width=0.923\linewidth, height=0.35\linewidth]{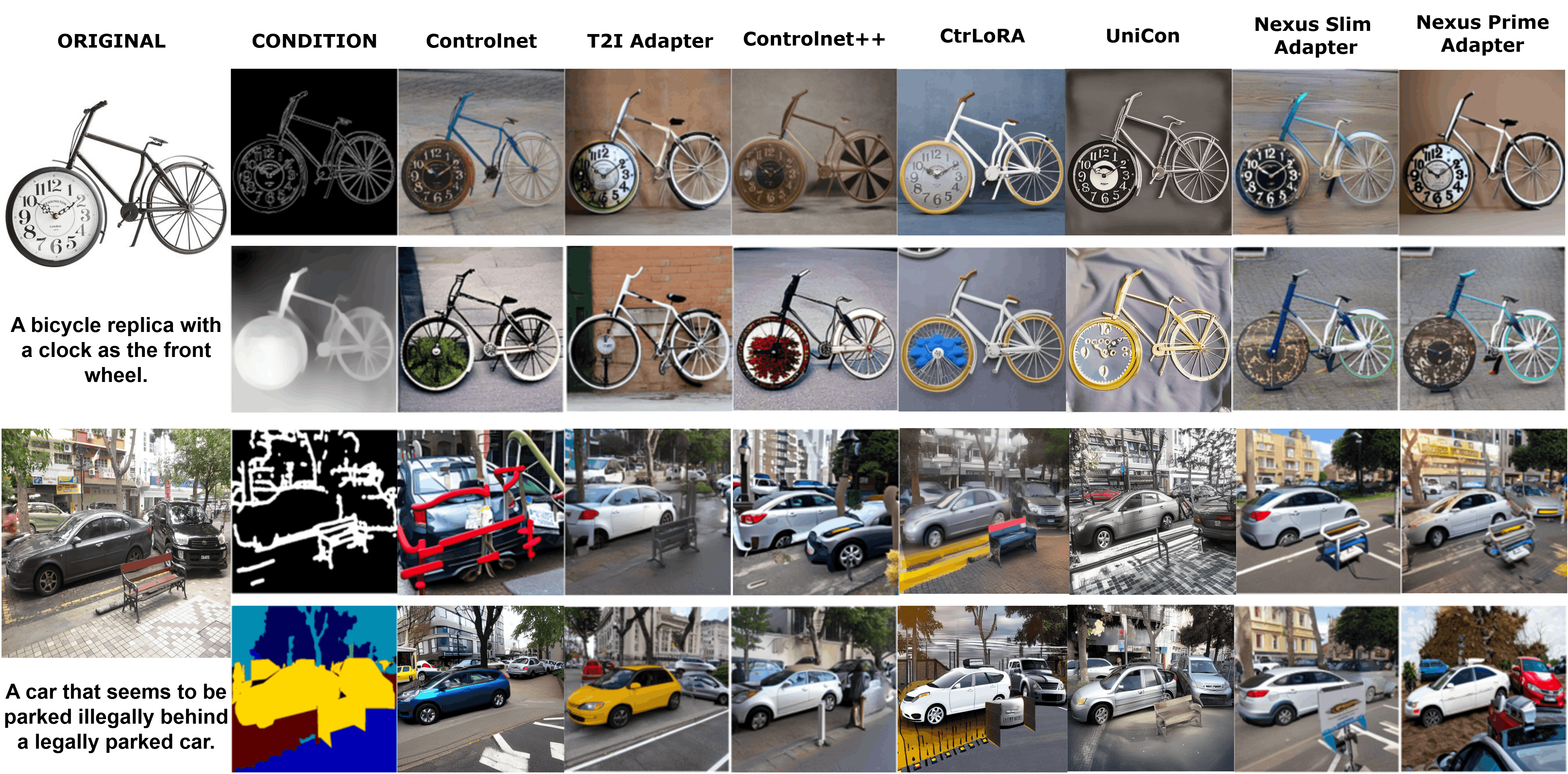}
    \vspace{-2mm}
    \caption{Qualitative comparison across four conditioning types showing that Nexus Prime consistently yields the most accurate and semantically aligned outputs, while Nexus Slim offers competitive performance with reduced complexity.}
    \label{fig:samples}
    \vspace{-5mm}
\end{figure*}

\begin{table}[t]
\centering
\scriptsize
\caption{Model complexity (↓) measured by Flops and number of trainable parameters. 
Lower values indicate higher efficiency. The best (lowest) values are in \textbf{bold}, 
the second-best are \underline{underlined}, and the third-best are \uwave{wavy underlined}.}
\vspace{-3mm}
\label{tab:params}
\vspace{2mm}
\resizebox{0.48\textwidth}{!}{%
\begin{tabular}{
>{\centering\arraybackslash}l|
>{\centering\arraybackslash}p{1.1cm}|
>{\centering\arraybackslash}p{1.8cm}|
>{\centering\arraybackslash}p{1.4cm}
}
\hline    
\textbf{Model}                     & \textbf{Venue} & \textbf{Flops (G)}             & \textbf{Params. (M)}       \\ \hline
ControlNet~\cite{controlnet}       & ICCV~2023      & 116.61                         & 361.28                     \\
T2I-Adapter~\cite{t2i}             & AAAI~2024      & \underline{29.97}              & \underline{77.37}          \\
ControlNet++~\cite{controlnet++}   & ECCV~2024      & 116.61                         & 361.28                     \\
\red{CtrLoRA}~\cite{ctrlora}       & ICLR~2025      & 116.61~$\xrightarrow{}$~135.15 & 361.28~$\xrightarrow{}$~37 \\
\red{UniCon}~\cite{unicon}         & ICLR~2025      & 111.62                         & 150                        \\ \hline
\textbf{Nexus Slim Adapter}        & --             & \textbf{23.77}                 & \textbf{59.29}             \\
\textbf{Nexus Prime Adapter}       & --             & \uwave{33.32}                  & \uwave{85.82}              \\ \hline
\end{tabular}
}
\vspace{-3mm}
\end{table}

\begin{table}[t]
\centering
\scriptsize
\setlength{\tabcolsep}{0.7mm}
\caption{Comparison of CLIP Scores (↑) across conditioning tasks. 
The highest values are shown in \textbf{bold}, the second-highest are \underline{underlined}, 
and the third-highest are \uwave{wavy underlined}.}
\vspace{-3mm}
\label{tab:clip_scores}
\vspace{2mm}
\resizebox{0.48\textwidth}{!}{%
\begin{tabular}{
        >{\centering\arraybackslash}l|
        >{\centering\arraybackslash}p{1cm}|
        >{\centering\arraybackslash}p{1cm}|
        >{\centering\arraybackslash}p{1cm}|
        >{\centering\arraybackslash}p{1.4cm}
    }
\hline
\textbf{Model}                     & \textbf{Canny}       & \textbf{Depth}         & \textbf{Sketch}        & \textbf{Segmentation} \\ \hline
ControlNet~\cite{controlnet}       & \underline{27.11}    & \uwave{27.21}          & \uwave{27.08}          & \uwave{26.74}         \\ 
T2I-Adapter~\cite{t2i}             & 26.42                & 26.16                  & 26.91                  & 26.69                 \\ 
ControlNet++~\cite{controlnet++}   & 26.79                & \underline{27.26}      & \underline{27.24}      & \textbf{27.19}        \\ 
\red{CtrLoRA}~\cite{ctrlora}       & \uwave{26.86}        & 26.00                  & 25.16                  & 25.32                 \\ 
\red{UniCon}~\cite{unicon}         & 26.62                & 27.09                  & 26.80                  & 26.71                 \\ \hline
\textbf{Nexus Slim Adapter}        & 26.39                & 26.71                  & 26.86                  & 25.89                 \\ 
\textbf{Nexus Prime Adapter}       & \textbf{27.33}       & \textbf{27.68}         & \textbf{27.66}         & \underline{27.03}     \\ \hline

\end{tabular}%
}
\vspace{-3mm}
\end{table}

\begin{table}[t]
\centering
\scriptsize
\setlength{\tabcolsep}{0.7mm}
\caption{Comparison of FID scores (↓) across different conditioning tasks. 
The lowest values are shown in \textbf{bold}, the second-lowest are \underline{underlined}, 
and the third-lowest are \uwave{wavy underlined}.}
\vspace{-3mm}
\label{tab:fid_scores}
\vspace{2mm}
\resizebox{0.48\textwidth}{!}{%
\begin{tabular}{
        >{\centering\arraybackslash}l|
        >{\centering\arraybackslash}p{1cm}|
        >{\centering\arraybackslash}p{1cm}|
        >{\centering\arraybackslash}p{1cm}|
        >{\centering\arraybackslash}p{1.4cm}
    }
\hline
\textbf{Model}                     & \textbf{Canny}       & \textbf{Depth}         & \textbf{Sketch}        & \textbf{Segmentation} \\ \hline
ControlNet~\cite{controlnet}       & 23.14                & 25.98                  & 25.23                  & 27.36                 \\ 
T2I-Adapter~\cite{t2i}             & 24.94                & 26.33                  & 26.81                  & 27.96                 \\ 
ControlNet++~\cite{controlnet++}   & 23.89                & \uwave{25.49}          & \underline{24.86}      & \textbf{25.63}        \\ 
\red{CtrLoRA}~\cite{ctrlora}       & \underline{22.89}    & 25.95                  & 26.32                  & 26.03                 \\ 
\red{UniCon}~\cite{unicon}         & 23.16                & 25.60                  & \uwave{24.91}          & 26.72                 \\ \hline
\textbf{Nexus Slim Adapter}        & \uwave{23.67}        & \underline{25.30}      & 26.40                  & \uwave{27.01}         \\ 
\textbf{Nexus Prime Adapter}       & \textbf{22.56}       & \textbf{23.91}         & \textbf{24.73}         & \underline{25.78}     \\ \hline

\end{tabular}%
}
\vspace{-5mm}
\end{table}

\red{Table~\ref{tab:params} compares the computational complexity of control-guided generative methods in terms of GFlops and parameters. ControlNet and ControlNet++ are the most resource-intensive, requiring 116.61 GFlops and 361.28M parameters, making them unsuitable for constrained environments. Both CtrLoRA and UniCon adopt LoRA-based fine-tuning strategies: CtrLoRA, when applied to ControlNet, reduces trainable parameters to 37M but still incurs high GPU memory usage since 361.28M frozen parameters remain active, increasing FLOPs (135.15). UniCon fine-tunes two UNet backbones with LoRA, adding to its computational cost. T2I-Adapter is comparatively efficient, requiring 29.97 GFlops and 77.37M total parameters. By contrast, the proposed Nexus Slim Adapter achieves the lowest complexity (23.77 GFlops, 59.29M total parameters), while Nexus Prime offers higher capacity (33.32 GFlops, 85.82M total parameters). Both Nexus variants demonstrate superior scalability, delivering high-quality conditional generation with reduced computational cost, making them practical for deployment in resource-limited environments.}

\red{Table~\ref{tab:clip_scores} reports CLIP Scores (↑) across Canny, Depth, Sketch, and Segmentation tasks. Nexus Prime Adapter achieves the highest scores on three tasks (Canny, Depth, Sketch) and ranks second on Segmentation, demonstrating strong text–structure fusion. ControlNet++ excels on Segmentation with the top score and consistently places second or third elsewhere, though at high computational cost. ControlNet remains competitive, ranking second on Canny and third on Depth and Sketch, effectively balancing structural cues with semantics. UniCon performs moderately, typically mid-tier. CtrLoRA underperforms overall, struggling particularly with Sketch (25.16) and Segmentation (25.32) due to overfitting to structural outlines at the expense of semantic detail. Nexus Slim scores lower than Prime but surpasses T2I-Adapter in most tasks, offering an appealing efficiency–performance trade-off. Overall, Nexus adapters deliver the best semantic–structural balance, with Prime leading and Slim excelling in efficiency.}

\red{Table~\ref{tab:fid_scores} presents FID scores (↓) across Canny, Depth, Sketch, and Segmentation. Nexus Prime attains the lowest FID in three tasks (Canny, Depth, Sketch) and ranks second in Segmentation, highlighting superior visual fidelity and consistency. ControlNet++ leads in Segmentation and ranks second on Sketch, reflecting strength in structure-intensive tasks. CtrLoRA unexpectedly secures second place on Canny but is moderate elsewhere, while ControlNet and UniCon remain solid mid-tier performers. Nexus Slim, though highly efficient, achieves second-best on Depth and remains competitive on Canny, but lags in Sketch and Segmentation. T2I-Adapter consistently records the weakest results across all tasks. Overall, Nexus Prime demonstrates the strongest balance of structural fidelity, semantic alignment, and image realism, while Nexus Slim offers an excellent efficiency–quality compromise.}

\red{While the Nexus Slim Adapter attains slightly lower CLIP scores than its Prime variant, it sustains strong FID results—ranking within the top three across tasks and outperforming several baselines. It surpasses the T2I-Adapter in both CLIP and FID with 23.36\% fewer parameters and lower FLOPs (Table~\ref{tab:params}), underscoring its efficiency–quality trade-off. In segmentation, ControlNet++ holds only a marginal edge over Nexus Prime, mainly from its feedback-based reward loss that regenerates conditions for richer region-specific learning, though this advantage is inconsistent. By contrast, CtrLoRA underperforms, as its low-rank updates bias heavily toward conditional inputs, limiting semantic alignment (Section~\ref{sec:ablation}). Crucially, Nexus adapters excel by fusing conditions with textual embeddings once—like T2I-Adapter—rather than repeatedly denoising them as in ControlNet, ControlNet++, or CtrLoRA. This unified embedding enables global guidance, improving semantic–structural consistency while lowering computational cost.}

\vspace{-1mm}
\subsection{Qualitative Visualization}

\begin{figure*}[t]
    \vspace{-4mm}
    \centering
    \includegraphics[width=0.95\linewidth, height=0.35\linewidth]{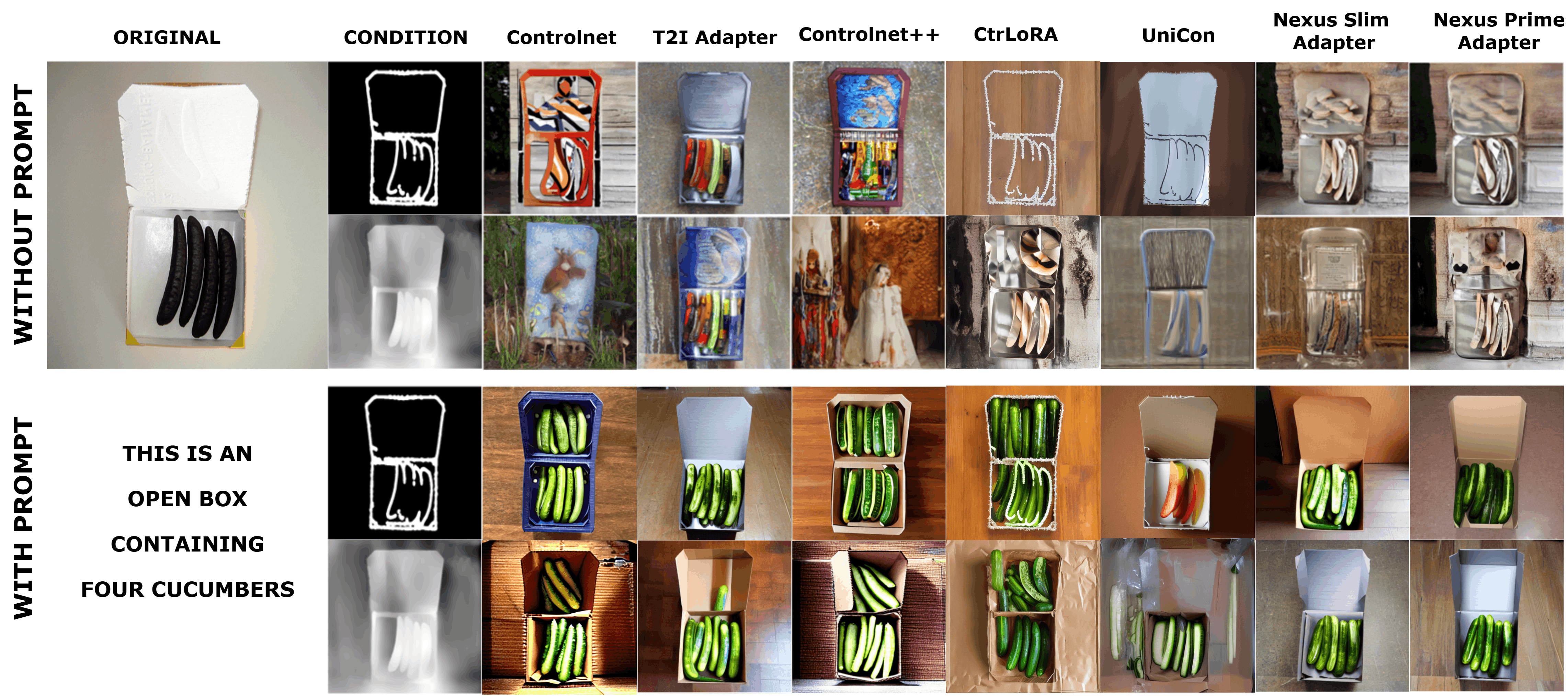}
    \vspace{-2mm}
    \caption{Qualitative comparison of image outputs with and without prompts, using only Sketch and Depth maps as conditional inputs.} 
    \label{fig:prompt_ablation}
    \vspace{-5mm}
\end{figure*}

In the Figure~\ref{fig:samples} we present a side-by-side qualitative comparison of generations conditioned on four structural input types: \textit{Canny edges} (first row), \textit{Depth} (second row), \textit{Sketch} (third row), and \textit{Segmentation} (fourth row). Each row visualizes the original reference image, its corresponding condition map, and the outputs generated baseline and proposed models. 

The first row shows the Canny-based generation, highlighting the models' ability to reconstruct fine edges. The Nexus Prime Adapter best preserves both geometric and semantic details, including the bicycle's structure and clock texture, due to its strong global guidance to the backbone. Nexus Slim performs well with slight texture loss. ControlNet variants deviate in realism, while T2I-Adapter retains semantics but lacks detailed fidelity. \red{CtrLoRA produces numbers, but the clock appears blurry and lacks detail.}
In the second row, we have shown the Depth conditioning enhances spatial realism and object placement. Nexus Prime delivers the most semantically accurate and spatially consistent results, effectively capturing vehicle arrangement and depth cues. Nexus Slim preserves structure with minor texture loss. ControlNet++ and ControlNet yield over-saturated or distorted outputs, while T2I-Adapter struggles with shape and depth consistency. \red{CtrLoRA correctly captures orientation but tends to overfit to the conditioning input, leading to biased generations, while UniCon produces structurally accurate outputs yet lacks overall realism.}
The third row shows the Sketch-based inputs evaluate outline reconstruction and interpretability. Nexus Prime generates coherent cars with accurate positioning from the sketch, due to its balanced global guidance. Nexus Slim captures the correct orientation but misses the second car. ControlNet and ControlNet++ fail to align with the sketch condition, while T2I renders a darker, less detailed version resembling Nexus Slim’s output. \red{CtrLoRA captures orientation but often produces blurry edges with reduced clarity, while UniCon aligns more closely with the prompt yet lacks fine details and overall visual coherence.}
The fourth row shows the Segmentation inputs demand semantic mapping and regional coherence. Nexus Prime excels with detailed, well-aligned car placements and realistic lighting. Nexus Slim also performs strongly, with slight simplifications. ControlNet++ and ControlNet show similar regional accuracy. T2I-Adapter struggles with consistent boundaries, often mixing semantic regions like overlapping vehicles or imprecise backgrounds. \red{CtrLoRA captures shape features from the condition but lacks realism and clarity, whereas UniCon produces more balanced outputs in this case, showing promise without evident bias.}

\noindent
\red{\textbf{Why do Nexus Adapters perform better?} 
Nexus adapters outperform existing methods through a fundamental architectural innovation. Unlike ControlNet-based approaches that operate as denoising partners alongside the backbone during inference—creating step-dependent guidance that can lead to inconsistent outputs. Nexus provides global, constant guidance throughout the generation process. This consistent flow enables the backbone to maintain coherent semantic-structural alignment across all denoising steps. ControlNet methods might suffer from over-reliance on step-wise adapter predictions, where structural conformity often overshadows semantic coherence, producing outputs that match conditions but lack contextual relevance to the prompt. The denoising-coupled approach creates dependency issues where misleading intermediate predictions can cascade through the generation process. In contrast, Nexus's prompt-guided global conditioning maintains constant semantic priors while providing structural cues, eliminating denoising dependency and enabling superior text-structure fusion. This architectural shift from step-wise conditioning to global guidance ensures both Nexus Prime and Slim achieve consistent improvements in fidelity, semantic, and structural coherence.}

\begin{table}[t]
\scriptsize
\centering
\small
\caption{Ablation on grouped convolution configuration in Nexus Slim Adapter Architecture (Canny condition).}
\vspace{-2.5mm}
\label{tab:group_ablation}
\vspace{1mm}
\begin{tabular}{c|c|c|c|c}
\hline
\textbf{Groups} & \textbf{FLOPs (G)} & \textbf{Params (M)} & \textbf{FID ↓} & \textbf{CLIP ↑} \\ \hline
4               & 17.36              & 42.23               & 27.14          & 25.11           \\
8               & 14.29              & 33.71               & 30.49          & 22.59           \\ \hline
\end{tabular}
\end{table}
\begin{table}[ht]
\scriptsize
\centering
\vspace{-2mm}
\caption{Ablation study comparing models under no-text-prompt settings using Sketch and Depth conditioning. FID ($\downarrow$) and CLIP ($\uparrow$) scores are reported. Two settings are evaluated: \textit{Without Prompt to Adapter (w/o P2A)} and \textit{Without Prompt to Stable Diffusion and Adapter (w/o P2SD+A)}.}
\vspace{-2mm}
\label{tab:no_prompt}
\resizebox{0.47\textwidth}{!}{%
\begin{tabular}{>{\centering\arraybackslash}p{10mm}|l|c|c|c|c}
\hline
\multirow{2}{*}{\textbf{Method}} 
& \multirow{2}{*}{\textbf{Model}} & \multicolumn{2}{c|}{\textbf{Sketch}} & \multicolumn{2}{c}{\textbf{Depth}} \\
\cline{3-6}
& & \textbf{FID} & \textbf{CLIP} & \textbf{FID} & \textbf{CLIP} \\ \hline

\multirow{5}{*}{\rotatebox{90}{\shortstack{\textbf{w/o} \\ \textbf{P2A}}}} 
& ControlNet           & 33.04 & 19.13 & 32.32 & 19.39 \\
& ControlNet++         & 32.56 & 19.24 & 31.98 & 19.31 \\
& \red{CtrLoRA}        & 30.41 & 20.18 & 30.17 & 20.22 \\
& \red{UniCon}         & 31.18 & 19.78 & 31.40 & 19.51 \\
& Nexus Slim Adapter   & \text{29.34} & \text{20.71} & \text{28.33} & \text{20.58} \\
& Nexus Prime Adapter  & \text{28.52}   & \text{21.63}  & \text{27.26}  & \text{20.94}  \\ \hline

\multirow{6}{*}{\rotatebox{90}{\shortstack{\textbf{w/o} \\ \textbf{P2SD+A}}}}
& ControlNet           & 38.84   & 17.29  & 37.81  & 16.22  \\
& T2I-Adapter          & \text{36.87} & \text{17.55} & \text{35.81} & \text{16.46} \\
& ControlNet++         & 39.20   & 17.00  & 38.13  & 15.98  \\
& \red{CtrLoRA}        & 35.66  & 16.89   & 34.85  & 16.48   \\
& \red{UniCon}         & 37.41 & 17.63 & 35.76 & 17.99 \\
& Nexus Slim Adapter   & \text{33.79} & \text{18.97} & \text{32.77} & \text{17.95} \\
& Nexus Prime Adapter  & \text{31.97}   & \text{19.77}  & \text{30.94}  & \text{18.73}  \\ \hline

\end{tabular}%
}
\vspace{-6mm}
\end{table}

\vspace{-1mm}
\subsection{Ablation Studies}\label{sec:ablation}

We conduct targeted ablations to analyze the efficiency-accuracy trade-offs and robustness of our method.

\textbf{Group Size Ablation:} 
To reduce parameters in the \textit{Nexus Slim Adapter}, we increase Grouped (\(G)\) \(3 \times 3\) convolutions to \(G=4\) and \(G=8\). While this improves efficiency by lowering FLOPs and parameter counts, it negatively impacts generation quality, especially at \(G=8\), where performance drops significantly, as shown in Table~\ref{tab:group_ablation}.

\red{\textbf{Prompt Guidance:}
We evaluate robustness without text prompts using Sketch and Depth conditioning (Figure~\ref{fig:prompt_ablation}). Without prompts, ControlNet-based methods exhibit catastrophic failure with severely distorted outputs, confirming their over-dependence on step-wise guidance during inference, while CtrLoRA maintains structural alignment but generates semantically incoherent artifacts and UniCon produces weakly aligned, blurred results. With prompt guidance, all methods improve significantly, though ControlNet variants still show recovery inconsistencies and CtrLoRA introduces unnatural artifacts. Nexus adapters demonstrate superior robustness across both conditions, consistently preserving structural fidelity and semantic coherence with minimal degradation, validating the effectiveness of global conditional guidance over inference-dependent approaches.
Table~\ref{tab:no_prompt} presents a validation study examining model stability in no-text conditions through two settings: Without Prompt to Adapter (w/o P2A), where only the adapter lacks textual input, and Without Prompt to Stable Diffusion and Adapter (w/o P2SD+A), where both components operate prompt-free. In w/o P2A, results highlight joint prompt-structure processing: Nexus Prime and Nexus Slim outperform others, confirming that conditional-focused alignment yields robust internal representations that preserve coherence without explicit prompts. Their efficient convolutional design with cross-attention maintains structural fidelity while using fewer parameters. CtrLoRA also performs strongly, indicating LoRA’s capacity to encode structural patterns when prompt dependencies are absent. UniCon achieves balanced outcomes across both settings, reflecting architectural stability approaching Nexus models. In w/o P2SD+A, ControlNet and ControlNet++ degrade sharply, validating their reliance on step-wise prompt guidance and exposing denoising-coupled fragility. By contrast, Nexus models degrade minimally, proving global conditional guidance ensures superior robustness. Overall, results demonstrate that conditional-focused alignment enables parameter-efficient, resilient architectures surpassing recent baselines.}


\vspace{-2mm}
\section{Conclusion}
In this work, we propose a pair of efficient and prompt-aware adapters for the diffusion model. The proposed adapter, Nexus Prime and Slim, helps to improve conditional image generation in the diffusion model. The earlier costly adapter, where they also consider the adapter to be independent of the prompt, preserves the conditional input but suffers in prompt alignment. The Nexus adapters integrate both structural inputs and textual prompts using cross-attention, enabling better semantic alignment without altering the frozen backbone. Our experimental results across various conditioning tasks demonstrate that Nexus Prime achieves superior fidelity, while Nexus Slim offers competitive performance with reduced computational cost. Quantitative and qualitative analyses confirm the effectiveness of the proposed model. Thus, the framework provides a solution for controllable, high-quality image generation.

{
    \small 
    \bibliographystyle{IEEEtran} 
    \bibliography{main.bib}
}

\clearpage
\setcounter{page}{1}
\maketitlesupplementary

\section{Additional Experiments}

\begin{table}[h]
\scriptsize
\centering
\vspace{-2mm}
\caption{Evaluation of models on the CUB-200 dataset for fine-grained image generation under \textit{Sketch} and \textit{Depth} conditions. FID ($\downarrow$) and CLIP ($\uparrow$) scores are reported under two configurations: \textit{Without Finetuning (w/o FT)} and \textit{With Finetuning (w FT)}.}
\vspace{-2mm}
\label{tab:cub}
\resizebox{0.47\textwidth}{!}{%
\begin{tabular}{>{\centering\arraybackslash}p{10mm}|l|c|c|c|c}
\hline
\multirow{2}{*}{\textbf{Method}} 
& \multirow{2}{*}{\textbf{Model}} & \multicolumn{2}{c|}{\textbf{Sketch}} & \multicolumn{2}{c}{\textbf{Depth}} \\
\cline{3-6}
& & \textbf{FID} & \textbf{CLIP} & \textbf{FID} & \textbf{CLIP} \\ \hline

\multirow{7}{*}{\rotatebox{90}{\shortstack{\textbf{\small w/o} \\ \textbf{\small FT}}}} 
& ControlNet~\cite{controlnet}        & 28.18 & 24.03 & 28.81 & 24.68 \\ 
& T2I-Adapter~\cite{t2i}              & 27.84 & 23.52 & 29.37 & 24.03 \\ 
& ControlNet++~\cite{controlnet++}    & 29.54 & 24.35 & 28.36 & 24.37 \\ 
& CtrLoRA~\cite{ctrlora}              & 28.83 & 23.03 & 28.80 & 22.76 \\ 
& UniCon~\cite{unicon}                & 27.73 & 24.41 & 28.79 & 24.13 \\ \cline{2-6}
& \textbf{Nexus Slim Adapter}         & 28.44 & 24.65 & 27.84 & 24.47 \\ 
& \textbf{Nexus Prime Adapter}        & 27.14 & 24.91 & 26.31 & 24.39 \\ \hline

\multirow{8}{*}{\rotatebox{90}{\shortstack{\textbf{\small w} \\ \textbf{\small FT}}}}
& ControlNet~\cite{controlnet}        & 26.25 & 26.48 & 27.29 & 26.62 \\ 
& T2I-Adapter~\cite{t2i}              & 25.88 & 26.14 & 27.55 & 23.93 \\ 
& ControlNet++~\cite{controlnet++}    & 27.47 & 26.88 & 26.70 & 24.37 \\ 
& CtrLoRA~\cite{ctrlora}              & 26.98 & 25.64 & 27.29 & 22.08 \\ 
& UniCon~\cite{unicon}                & 25.96 & 26.21 & 26.19 & 23.60 \\ \cline{2-6}
& \textbf{Nexus Slim Adapter}         & 26.81 & 26.14 & 25.79 & 24.15 \\ 
& \textbf{Nexus Prime Adapter}        & 25.72 & 26.69 & 25.25 & 24.46 \\ \hline

\end{tabular}%
}
\vspace{-3mm}
\end{table}

To validate whether the proposed Nexus Adapters can handle complex image generation in fine-grained tasks, we conduct two evaluations. First, we test our model on the CUB-200 dataset~\cite{cub200} and compare it against baselines. Second, we fine-tune all models for $1k$ steps to measure adaptability and efficacy. Prior work, such as CtrLoRA~\cite{ctrlora}, suggests that ControlNet with a LoRA-based architecture adapts quickly with limited data (1k images). However, our observations show that fidelity often degrades in complex generations, with noticeable loss of color and spatial consistency. We fine-tune all architectures under two distinct conditions—\textit{Sketch} and \textit{Depth}—and report results in Table~\ref{tab:cub}.

The results highlight that the adaptation capabilities of the proposed Nexus Adapters are largely on par with later state-of-the-art models such as ControlNet++~\cite{controlnet++}, CtrLoRA, and UniCon. Without fine-tuning, most models face challenges in maintaining detail and consistency, but fine-tuning consistently improves performance across both \textit{Sketch} and \textit{Depth} conditions. Nexus Slim and Prime Adapters demonstrate strong competitiveness, narrowing the gap with heavier baselines while retaining their lightweight design advantages. These observations suggest that adapter-based approaches can match the adaptability of established architectures while offering a more efficient path for fine-grained image generation tasks.

\section{Training and Inference Details}

\begin{table}[t]
\small
\centering
\caption{Estimated training and inference time across models. Training time is computed for $200k$ steps with batch size 2.}
\vspace{-2mm}
\label{tab:time_estimates}
\resizebox{\linewidth}{!}{
\begin{tabular}{l|c|c}
\hline
\textbf{Model} & \textbf{Train Time (hrs)} & \textbf{Infer Time (ms/img)} \\ \hline
ControlNet              & $\approx$~129  & $\approx$~38 \\
T2I-Adapter             & $\approx$~33   & $\approx$~9  \\
ControlNet++            & $\approx$~129  & $\approx$~38 \\
CtrLoRA                 & $\approx$~150  & $\approx$~45 \\
UniCon                  & $\approx$~124  & $\approx$~37 \\\hline
Nexus Slim Adapter      & $\approx$~26   & $\approx$~7  \\
Nexus Prime Adapter     & $\approx$~37   & $\approx$~11 \\ \hline
\end{tabular}
}
\vspace{-5mm}
\end{table}

Both Nexus Prime and Nexus Slim architectures are trained for 200,000 steps at an image resolution of $512 \times 512$ using mixed-precision (\textit{fp16}) training, which significantly reduces memory footprint and speeds up computation without degrading performance. A batch size of 2 is employed, with gradient accumulation steps set to 4, effectively simulating a batch size of 8 to stabilize training under GPU memory constraints. The optimizer is AdamW~\cite{adamw}, configured with a constant learning rate of $5 \times 10^{-6}$, $\beta_1 = 0.9$, $\beta_2 = 0.999$, $\epsilon = 1 \times 10^{-8}$, and a weight decay of $1 \times 10^{-2}$. A short linear warm-up of 500 steps ensures stability in the early training phase, after which a constant learning rate schedule is maintained. 

During inference, we adopt a standard classifier-free guidance strategy with a guidance scale of $7.5$, which balances fidelity to the text prompt against structural adherence to the conditioning input. Sampling is performed with 35 denoising steps, offering a practical trade-off between image quality and computational efficiency. 
All experiments were conducted on a high-performance server equipped with an \textit{AMD EPYC 7763 64-Core} CPU, 128 GB of RAM, and \textit{2 NVIDIA A100 80 GB} GPUs, running \textit{CUDA 12.2}. This configuration enables efficient large-scale training and reliable evaluation of the proposed architectures. 

Table~\ref{tab:time_estimates} summarizes the estimated training and inference costs across models, measured over 200K steps (batch size 2) and average per-image latency on a single A100 GPU. ControlNet and ControlNet++ are the most resource-intensive, each requiring $\sim$129 hours of training and 38 ms per image. CtrLoRA is even heavier, at 150 hours and 45 ms, due to the overhead of low-rank updates. UniCon slightly reduces cost (124 hrs, 37 ms), while T2I-Adapter is far more efficient, training in only 33 hours with 9 ms inference. Nexus Slim achieves the best efficiency (26 hrs, 7 ms), and Nexus Prime balances performance with moderate cost (37 hrs, 11 ms). For fairness, inference is reported with 35 steps across all models. 


\section{Additional Ablations}

\begin{table}[t]
\centering
\scriptsize
\setlength{\tabcolsep}{1.5mm}
\caption{Ablation on the number of adapter blocks. FID (↓) and CLIP Score (↑) for Depth and Sketch conditioning tasks. The 4-block configuration represents the full model.}
\vspace{-2mm}
\label{tab:blocks_ablation}
\resizebox{0.48\textwidth}{!}{%
\begin{tabular}{
        >{\centering\arraybackslash}c|
        >{\centering\arraybackslash}c|
        >{\centering\arraybackslash}c
        >{\centering\arraybackslash}c|
        >{\centering\arraybackslash}c
        >{\centering\arraybackslash}c
    }
\hline
\multirow{2}{*}{\textbf{Blocks}} & \multirow{2}{*}{\textbf{Model}} & \multicolumn{2}{c|}{\textbf{Depth}} & \multicolumn{2}{c}{\textbf{Sketch}} \\ \cline{3-6} 
 & & \textbf{FID↓} & \textbf{CLIP↑} & \textbf{FID↓} & \textbf{CLIP↑} \\ \hline
2               & Nexus Slim Adapter           & 33.43          & 17.69          & 32.88          & 18.35 \\
                & Nexus Prime Adapter          & 28.52          & 21.98          & 30.50          & 20.49 \\ \hline
3               & Nexus Slim Adapter           & 30.38          & 20.75          & 29.95          & 20.86 \\
                & Nexus Prime Adapter          & 26.11          & 22.30          & 27.42          & 21.53 \\ \hline
\textbf{4}      & \textbf{Nexus Slim Adapter}  & \textbf{25.30} & \textbf{26.71} & \textbf{26.40} & \textbf{26.86} \\
\textbf{(ours)} & \textbf{Nexus Prime Adapter} & \textbf{23.91} & \textbf{27.68} & \textbf{24.73} & \textbf{27.66} \\ \hline
\end{tabular}%
}
\vspace{-5mm}
\end{table}

\begin{figure*}[t]
    \centering
    \includegraphics[width=1\linewidth]{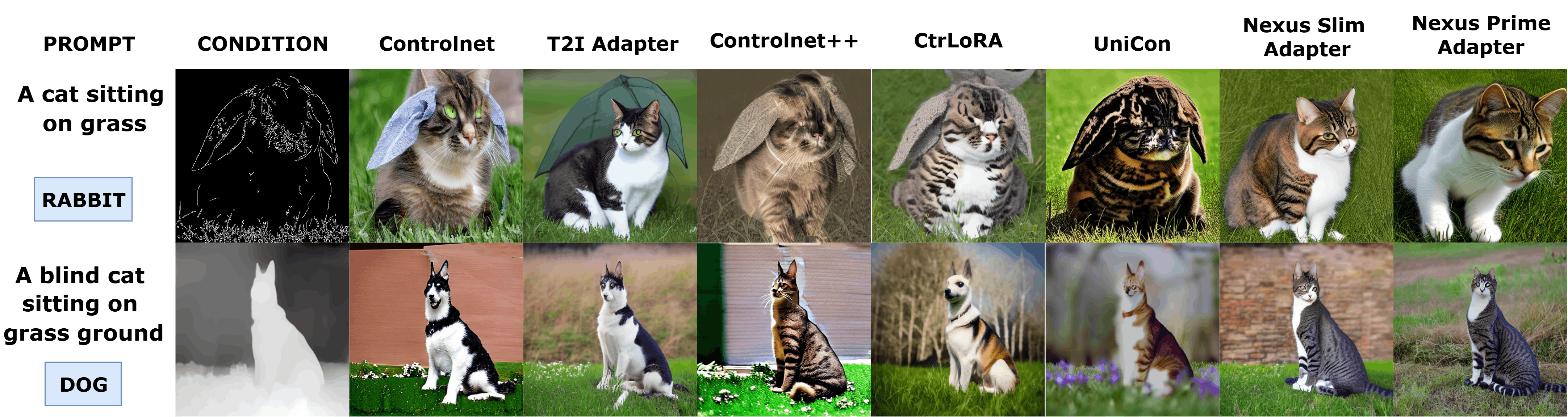}
    \caption{Qualitative ablation on conflicting prompts and conditional images. The actual object in the image-conditioning input is highlighted with a blue background, and a conflicting prompt is provided as test input. We can observe that the model preserve the image-conditioned structural input while generating the object based on the text prompt.}
    \label{fig:contradiction}
\end{figure*}

\textbf{Number of Adapter Blocks:} To evaluate whether adapter blocks in deeper UNet layers remain necessary—where structural information may diminish and the model relies more on the text prompt—we perform an ablation varying the number of adapter blocks (2, 3, and 4) in both Nexus Slim and Nexus Prime Adapters. Table~\ref{tab:blocks_ablation} reports FID (↓) and CLIP Score (↑) for Depth and Sketch conditioning. Increasing the number of blocks consistently improves both FID and CLIP scores, with the full 4-block configuration (ours) achieving the best performance. This demonstrates that deeper adapter blocks effectively preserve fine-grained structural details while still integrating prompt guidance, validating that the final layers contribute meaningful improvements rather than being redundant. These results confirm that careful design of adapter depth is critical for maintaining structural fidelity and semantic alignment in conditional image generation.

\textbf{Conflict through Prompt:} In the Figure~\ref{fig:contradiction} we presents a qualitative ablation on conflicting prompts and conditional images, where the actual object in the conditioning input is highlighted in blue. This study validates that incorporating textual information in the conditioning is crucial for generating coherent images with proper alignment to both shape and structure, alongside accurate subject mapping. The Nexus Prime Adapter consistently produces outputs that respect the conditioning structure while adapting to the prompt, yielding coherent and realistic results. Nexus Slim preserves structural fidelity but slightly sacrifices fine details. ControlNet and ControlNet++ often overfit to the condition, ignoring the prompt and producing semantically inconsistent outputs. CtrLoRA captures basic structure but introduces blurring and artifacts, while UniCon generates plausible objects yet struggles with realism under conflicting prompts. Overall, these results emphasize the superior balance of prompt and condition integration achieved by the Nexus adapters compared to existing baselines.

\end{document}